\documentclass[letterpaper, 10 pt, journal, twoside]{IEEEtran}
\IEEEoverridecommandlockouts
 
\usepackage{cite}
\usepackage{amsmath,amssymb,amsfonts}
\usepackage{graphicx}
\usepackage{textcomp}
\usepackage{color}
\usepackage{subfigure}
\usepackage{multirow}
\usepackage{multicol}
\usepackage{makecell}
\usepackage{bm}
\usepackage{algorithm}
\usepackage{algpseudocode}
\usepackage{hyperref}
\newcommand{\ie}{i.e.,\ }
\newcommand{\eg}{e.g.,\ }
\newcommand{\Reffig}[1]{Figure~\ref{#1}}
\newcommand{\Refsec}[1]{Section~\ref{#1}}

\newcommand{\Refeq}[1]{Equation~\eqref{#1}}
\newcommand{\Reftab}[1]{Table~\ref{#1}}

\usepackage{graphics} 
\usepackage{epsfig} 
\usepackage{algorithm}
\usepackage{algpseudocode}
\usepackage{amsmath}
\usepackage{amssymb}
\usepackage{subfigure}
\usepackage{graphicx}
\usepackage{subfigure}
\usepackage{booktabs}
\usepackage{threeparttable}
\usepackage{multirow}
\usepackage{multicol}
\usepackage{makecell}
\usepackage{dcolumn}
\usepackage[T1]{fontenc}
\usepackage{diagbox}

\begin{document}
\title{
        Learning Sequence Descriptor based on Spatio-Temporal Attention for Visual Place Recognition
}

\author{Junqiao Zhao$^{1 , 2}$, Fenglin Zhang$^{*1}$, Yingfeng Cai$^{1}$, Gengxuan Tian$^{1}$, Wenjie Mu$^{1}$, Chen Ye$^{1}$, Tiantian Feng$^{3}$ 
\thanks{Manuscript received: October 12, 2023; Revised December 20, 2023; Accepted December 31, 2023.}
\thanks{
This paper was recommended for publication by Editor Sven Behnke upon evaluation of the Associate Editor and Reviewers' comments.
This work is supported by the National Key Research and Development Program of China (No. 2021YFB2501104). \emph{(Corresponding Author: Fenglin Zhang)}} 

\thanks{$^{1}$Department of Computer Science and Technology, 
School of Electronics and Information Engineering, Tongji University, Shanghai, China, and the MOE Key Lab of Embedded System and Service Computing, Tongji University, Shanghai, China}
\thanks{$^{2}$Institute of Intelligent Vehicles, Tongji University, Shanghai, China}
\thanks{$^{3}$School of Surveying and Geo-Informatics, Tongji University, Shanghai, China}
\thanks{Digital Object Identifier (DOI): see top of this page.}
}

\markboth{IEEE Robotics and Automation Letters. Preprint Version. Accepted January, 2024}
{Zhao \MakeLowercase{\textit{et al.}}: Learning Sequence Descriptor based on Spatio-Temporal Attention for Visual Place Recognition} 

\maketitle

\begin{abstract}
Visual Place Recognition (VPR) aims to retrieve frames from a geotagged database that are located at the same place as the query frame. To improve the robustness of VPR in perceptually aliasing scenarios, sequence-based VPR methods are proposed. These methods are either based on matching between frame sequences or extracting sequence descriptors for direct retrieval. However, the former is usually based on the assumption of constant velocity, which is difficult to hold in practice, and is computationally expensive and subject to sequence length. Although the latter overcomes these problems, existing sequence descriptors are constructed by aggregating features of multiple frames only, without interaction on temporal information, and thus cannot obtain descriptors with spatio-temporal discrimination.
In this paper, we propose a sequence descriptor that effectively incorporates spatio-temporal information. Specifically, spatial attention within the same frame is utilized to learn spatial feature patterns, while attention in corresponding local regions of different frames is utilized to learn the persistence or change of features over time. We use a sliding window to control the temporal range of attention and use relative positional encoding to construct sequential relationships between different features. This allows our descriptors to capture the intrinsic dynamics in a sequence of frames.
Comprehensive experiments on challenging benchmark datasets show that the proposed approach outperforms recent state-of-the-art methods.
The code is available at \url{https://github.com/tiev-tongji/Spatio-Temporal-SeqVPR}.
\end{abstract}


\IEEEpeerreviewmaketitle

\section{INTRODUCTION}

\IEEEPARstart{V}{isual} place recognition (VPR) aims to retrieve frames from a geotagged database that are located at the same place as the queried frame \cite{lowry2015survey}.
It is typically used for loop detection in simultaneous localization and mapping (SLAM) as well as for visual relocalization.
Various approaches have been proposed to enhance the performance of VPR by learning improved single frame representation \cite{vlad, fishervector, bow, chen2014convolutional, arandjelovic2016netvlad, jin2017crn}.
However, single frame-based VPR is vulnerable to drastic changes in viewpoint and appearance, so studies have delved into the utilization of sequence information to address this issue.

One category of sequence-based methods is based on sequence matching.
This approach involves comparing each frame of the query sequence with the database to create a matching matrix.
Then, the diagonal values are aggregated to obtain a similarity score to determine the location of the query sequence.
However, this method is mainly suitable for cases where the camera motion remains relatively stable \cite{milford2012seqslam}.
Otherwise, incorrect matches may occur.
Additionally, the computational cost of sequence matching increases with the sequence length and map size \cite{siam2017fastseqslam}.

To overcome the aforementioned challenges, researchers have proposed utilizing descriptors to represent a sequence \cite{facil2019condition}.
Sequence descriptors offer better scalability for varying sequence lengths and greater robustness against perceptual aliasing.
However, existing research only aggregates descriptors \cite{garg2021seqnet} or local features of multiple frames \cite{mereu2022seqvlad}, neglecting the cross-frame temporal interactions, which makes sequence descriptors less discriminative.

In this paper, we propose an approach for exploring spatio-temporal interactions within frame sequences to extract sequence descriptors. 
Such sequence descriptors take into account both the temporal correlation across multiple frames and the spatial structure distribution in a frame.
By employing the attention mechanism, we adaptively weight image patches to capture and combine discriminative features in the sequence.
A sliding window is used to control the attentional range and reduce the computational burden.
Moreover, relative positional encoding is employed to guide the sequence descriptors in learning spatio-temporal patterns rather than specific visual content.
This choice stems from the observation that, during camera motion, the visual content moves with the frame, while the relative positions of spatio-temporal patterns remain constant.

The contributions of this paper are threefold:
\begin{itemize}
\item We introduce a spatio-temporal sequence descriptor that effectively captures the interaction of the spatial and temporal information simultaneously.
\item We investigate the impact of positional encoding on the spatial and temporal information interactions.
\item Our approach delivers competitive results across multiple datasets, outperforming existing state-of-the-art methods based on sequence descriptors.
\end{itemize}


\section{RELATED WORKS}

\subsection{Sequence-based VPR}

There are mainly two avenues for utilizing sequence information in VPR: sequence matching and sequence descriptor extraction \cite{lowry2015survey}.

Sequence matching involves two key steps.
Initially, a similarity matrix is constructed by comparing the descriptors of each frame in the query sequence with the descriptors of all frames in the database.
Subsequently, the most similar sequence in the database is determined by aggregating the individual similarity scores. 
SeqSLAM \cite{milford2012seqslam} is a pioneering example of sequence matching.
However, SeqSLAM can be computationally demanding especially when handling large maps.
Additionally, it relies on the assumption of constant velocity, which can limit its applicability in scenarios with varying motion characteristics. 
To address these challenges, several innovations have emerged.
Fast-SeqSLAM \cite{siam2017fastseqslam} leverages an approximate nearest neighbor algorithm to reduce time complexity without degrading accuracy.
\cite{lu2019dtw} proposes a local matching method based on an improved dynamic time warping algorithm, which relaxes the assumption of constant velocity and concurrently reduces time complexity.
\cite{naseer2014robust} and \cite{vysotska2019effective} use a cost matrix-based approach via dynamic programming to alleviate the issues of missing frames.
These methods have also been evaluated on sequences with strong seasonal changes and showing promising performance.
However, these methods operate on the matching scores obtained from the underlying single frame descriptors.

In the second avenue, a descriptor is extracted to represent a sequence, followed by a direct sequence-to-sequence similarity search.
This not only reduces the matching cost but also incorporates temporal cues into the descriptor.
\cite{facil2019condition} first proposes the idea of fusing multiple individual descriptors to generate a sequence descriptor.
Subsequently, SeqNet \cite{garg2021seqnet} proposes using a 1-D convolution to learn frame-level features into a sequence descriptor.
However, this approach is implemented based on pre-computed individual frame descriptors, which prevents the sequence descriptors from capturing the local features within each frame.
SeqVLAD \cite{mereu2022seqvlad} proposes a detailed taxonomy of techniques using sequence descriptors.
It analyzes various mechanisms for fusing individual frame information, and further investigates the feasibility of using the Transformer as the backbone. 
The sequence descriptor is aggregated by NetVLAD \cite{arandjelovic2016netvlad} directly from the local features of each frame in the sequence.
However, it does not consider the temporal information interaction across frames, resulting in descriptors without spatio-temporal discrimination.

\subsection{Spatio-temporal Attention Mechanism}

Spatio-temporal attention mechanisms have been applied in various tasks, including video retrieval, video classification, and more. 
In the context of video action recognition, \cite{feichtenhofer2017spatiotemporal} presents a general ConvNet architecture. 
It leverages multiplicative interactions of spatio-temporal features to capture long-term dependencies among local features.
\cite{li2020spatio} proposes a spatio-temporal attention network to learn discriminative feature representations for actions.
In the video classification task, \cite{bertasius2021space} explores the efficacy of spatio-temporal attention mechanism for feature learning directly from image patches.
In the realm of video action recognition and object detection,
\cite{li2022mvitv2} introduces a novel multi-scale vision transformer, which achieves state-of-the-art performance.
The spatio-temporal attention mechanism has also been extended to diverse tasks such as image captioning \cite{ji2020spatio} and person re-identification \cite{aich2021spatio}. 

These methods leverage both spatial and temporal information to selectively focus on relevant video regions or frames.
Interestingly, despite the success of spatio-temporal attention mechanism in various applications, it has not yet been integrated into sequence-based VPR.

\section{METHODOLOGY}
We begin by presenting the architecture, which encompasses spatio-temporal-based feature learning and aggregation.
Subsequently, we introduce the loss function.
\begin{figure*}
    \centering
    \includegraphics[width=18cm]{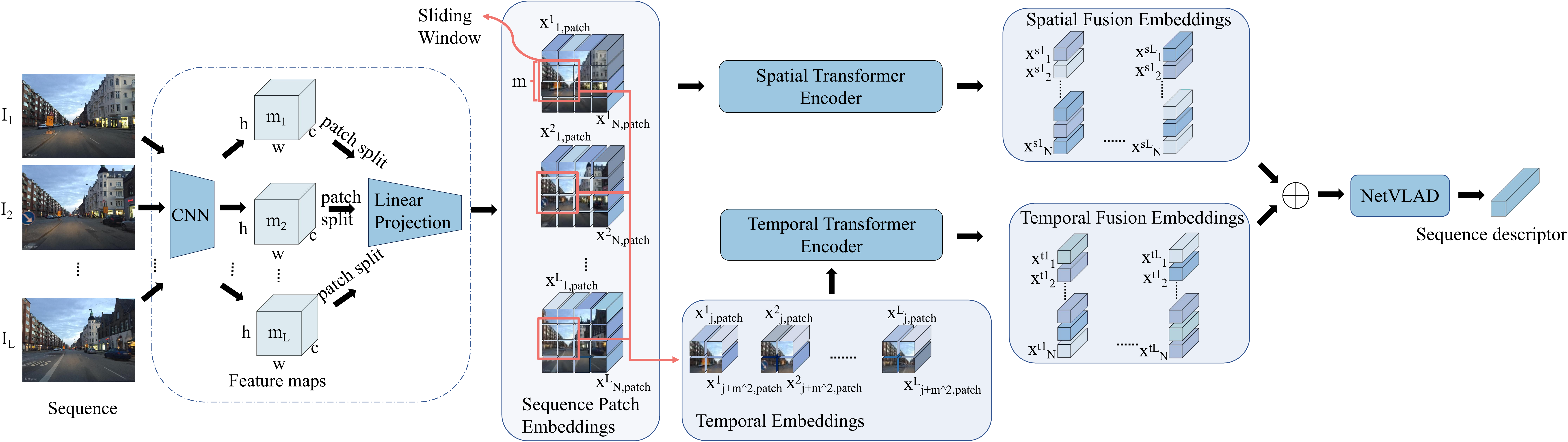}
    \caption{\textbf{The architecture of our proposed method.} 
    Given a continuous sequence of raw frames $I_{1}$, $I_{2}$, $\ldots$, $I_{L}$, we employ a Convolutional Neural Network (CNN) to map each frame to feature maps and then split these maps into patches.
    A Linear Projection is subsequently employed to map the patch features to embeddings $\{x^{1}_{1,\ldots,N},x^{2}_{1,\ldots,N},\ldots,x^{L}_{1,\ldots,N}\}$.
    These embeddings from individual frames are then passed through Spatial Transformer Encoders, applying self-attention for spatial information interaction.
    This process yields a set of transformed embeddings $\{x^{s1}_{1,\ldots,N},\ldots,x^{sL}_{1,\ldots,N}\}$.
    Furthermore, the embeddings across different frames within the sliding window are input into a Temporal Transformer Encoder to fuse temporal information, which generates $\{x^{t1}_{1,\ldots,N},\ldots,x^{tL}_{1,\ldots,N}\}$.
    Finally, the embeddings from two branches are combined, and the NetVLAD layer is employed to aggregate these embeddings to generate a sequence descriptor.
    }
    \label{pipeline}
\end{figure*}

\subsection{Architecture}

The architecture of our model is illustrated in \Reffig{pipeline}.
It takes a frame sequence $S\in\mathbb{R}^{L\times H\times W\times 3}$ as input, composed of $L$ image frames with dimensions $H\times W$.
Since vision transformer (ViT) \cite{dosovitskiy2021anvit} is computation-intensive and lacks the inductive biases inherent in convolutional neural networks (CNN) \cite{hassani2021cct}, we utilize convolution layers to map each frame $s_{i}$ to a feature map $m_{i}\in\mathbb{R}^{h\times w\times c}$, where $s_{i}\in S$, $i\in\{1,2,\ldots,L\}$ and $c$ represents the number of channels.
Subsequently, the feature map of each frame is split into $N$ non-overlapping patches, where $N$ is determined as $hw/P^{2}$, given the patch size of $P\times P$.
Following this, each patch is flattened and mapped to an embedding $x$ using a trainable linear projection $E$ as \Refeq{patchembeding}.

\begin{equation}
    x_{j,patch}^{i} = E(p_{j}^{i}),\ x_{j}^{i}\in \mathbb{R}^{D}
    \label{patchembeding}
\end{equation}
where $i\in\{1,2,\ldots,L\}$, $j\in\{1,2,\ldots,N\}$, and $p \in \mathbb{R}^{P \times P}$ which indicates the patch.
Then, $x$ is employed as the input embedding for Spatial and Temporal Transformer Encoders.

\subsubsection{Spatial Attention}
In each frame, the positions of the local features reflect the spatial distribution of the features, and this distribution remains relatively consistent within a frame under the same view.
Similar to ViT \cite{dosovitskiy2021anvit} and as illustrated in \Reffig{position} (a), we incorporate position information into the patch embedding using a standard learnable absolute positional embedding denoted as $x_{pos}\in \mathbb{R}^{D}$, as follows\footnote{The superscript is omitted where it does not cause ambiguity.}:
\begin{equation}
    x_{j} = x_{j,patch} + x_{j,pos},\ x_{j}\in \mathbb{R}^{D}
    \label{posembeding}
\end{equation}

We employ an $L_{s}$-layer transformer encoder for spatial fusion, which outputs spatial fusion embeddings $\{x^{s1}_{1,\ldots,N},\ldots,x^{sL}_{1,\ldots,N}\}$.
Each layer consists of a multi-head self-attention (MSA) module, a multi-layer perceptron (MLP), Layer-norm (LN) blocks and residual connections.
In the MSA module, linear projections $W^{h}_{Q}$, $W^{h}_{K}$, $W^{h}_{V}$$\in\mathbb{R}^{d\times D}$ are applied to query ($Q^h$), key ($K^h$) and value ($V^h$) according to \Refeq{qkv}, where $X=\{x_{1},x_{2},\ldots,x_{N}\}$, $h$ represents the head index and $d=D/h$.
Subsequently, the self-attention weights are computed through the dot-product of $Q^{h}$ and $K^{h}$, then the output $Z^{h}\in\mathbb{R}^{N\times d}$ is generated by multiplying scaled weights and $V^{h}$ in \Refeq{eqattention}.
Finally, the output $\{Z^{1},Z^{2},\ldots, Z^{h}\}$ from the heads are concatenated to form $Z\in\mathbb{R}^{N\times D}$ in \Refeq{concat}, which serves as input to the Layer-norm and MLP components.

\begin{equation}
    Q^{h} = W^{h}_{Q}X,\ K^{h} = W^{h}_{K}X,\ V^{h} = W^{h}_{V}X
    \label{qkv}
\end{equation}
\begin{equation}
    Z^{h} = \text{Softmax}(Q^{h}{K^{h}}^{T}/\sqrt{d})V^{h}
    \label{eqattention}
\end{equation}
\begin{equation}
    Z = \text{Concat}(Z^{1},Z^{2},\ldots, Z^{h})
    \label{concat}
\end{equation}

\subsubsection{Temporal Attention}
\begin{figure*}
    \centering
    \includegraphics[width=16cm]{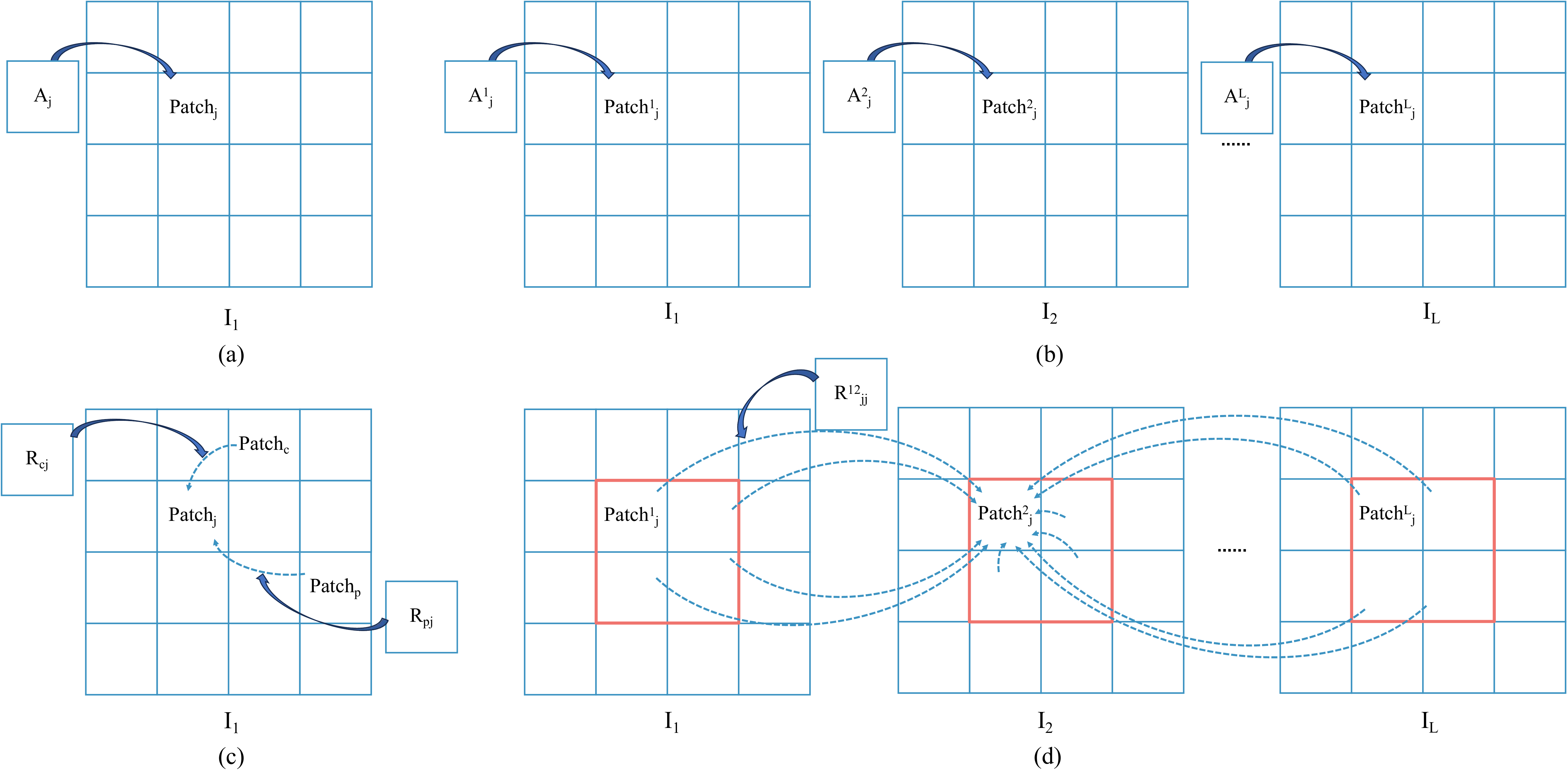}
    \caption{\textbf{The different positional embeddings and sliding windows.} 
    Here we show spatial absolute positional embedding (a) and relative positional embedding (c), as well as temporal absolute positional embedding (b) and relative positional embedding (d),
    where we use $A$ to represent the absolute positional embedding and $R$ for the relative positional embedding.
    Dashed arrows indicate that information is passed between the two patches.
    Solid arrows indicate to fuse positional embeddings to the information passing.
    It's important to note that absolute positional embeddings are independent of the inter-patch relationships. 
    In contrast, relative positional embeddings vary based on the position relationship between patches.
    }
    \label{position}
\end{figure*}

We define a sliding window of size $m\times m$ to control the temporal self-attention range in the sequence.
Within each layer of the temporal transformer attention, self-attention is performed within identical sliding windows of the same region across multiple images.
The window moves along the rows and columns, indicating that attention is performed between temporally adjacent regions, as illustrated in \Reffig{position} (d), rather than between two images.
In the temporal interaction, $x_{j,patch}$ from \Refeq{patchembeding} is taken as the input.
We redefine $X$ in \Refeq{qkv} as $X=\{x^{1}_{j},\ldots,x^{1}_{j+m^{2}},\ldots,x^{L}_{j},\ldots,x^{L}_{j+m^{2}}\}$, where $L$ is the frame index, $j$ is the patch index and $m$ is the sliding window size, and we generate $Q$, $K$, $V$ respectively.
Compared to absolute positions, we argue that relative positions provide more accurate description of the consistency or variation of local features in a sequence over time. 
This is because relative position information can capture the changing relationship between the positions of two patches across different frames, as illustrated in \Reffig{position} (d), whereas absolute position information merely considers the static relationship between one patch and all patches, as illustrated in \Reffig{position} (b).

We encode the relative position between two input embeddings $x_i$ and $x_j$ in $X$, into a relative positional embedding $P_{ij}\in \mathbb{R}^{D}$, following \cite{shaw2018self}.
The representation of pairwise encoding is then embedded into the self-attention module\footnote{Here we only show the single-head self-attention, for multi-head self-attention, please refer to \Refeq{qkv}, \Refeq{eqattention} and \Refeq{concat}.},
\begin{equation}
    Z = \text{Softmax}(QK^{T}+E^{(rel)}/\sqrt{D})V
    \label{relattention}
\end{equation}
where $E^{(rel)}_{ij}=Q_{i}P_{ij}$, and $E^{(rel)} \in \mathbb{R}^{(m\times m\times L)\times(m\times m\times L)}$, and the $E$ changes in each layer of the temporal transformer encoder.
We further decompose the relative positional embedding into height, width and temporal axes following \cite{li2022mvitv2}, which can reduce the number of learnable parameters.
We adopt the $L_{t}$-layer transformer encoder for temporal fusion, then the temporal fusion embeddings $\{x^{t1}_{1,\ldots,N},\ldots,x^{tL}_{1,\ldots,N}\}$ are generated.

\subsubsection{Aggregation} \label{sec_aggregation}
In our spatial and temporal attention blocks, the class token is removed.
This decision aligns with our strategy of utilizing attention for information interaction, rather than extracting frame descriptors.
The sequence descriptor is aggregated by NetVLAD \cite{arandjelovic2016netvlad}.

Given a set of D-dimensional embeddings from spatial and temporal Transformer, we combine them based on the position of patches \ie $\{x^{s1}_{1,\ldots,N}+x^{t1}_{1,\ldots,N},\ldots,x^{sL}_{1,\ldots,N}+x^{tL}_{1,\ldots,N}\}$.
Following this, we perform aggregation on the resulting $N\times L$ embeddings using NetVLAD, 
\begin{equation}
    V(k) = \sum _{i=1}^{N\times L}a_{k}(x_{i})(x_{i}-c_{k})
    \label{netvlad}
\end{equation}
where $x_{i}$ is a single embedding, $c_{k}$ is the $k$-th centroid which is trainable parameter, and the $a_{k}(x_{i})$ is a soft-assignment defined as:
\begin{equation}
    a_{k}(x_{i}) = \frac{e^{w_{k}^{T}x_{i}+b_{k}}}{\sum _{k'}e^{w_{k'}^{T}x_{i}+b_{k'}}} 
    \label{softassignment}
\end{equation}
where $w_{k}$, $b_{k}$ are also trainable parameters.

\subsection{Loss Function}
Similar to the training regime of NetVLAD, we use the max-margin triplet loss as below:
\begin{equation}
    Loss =  \sum_{k}  \max (\left\lVert S_{a} - S_{p}\right\rVert _{2} - \left\lVert S_{a} - S^{k}_{n}\right\rVert _{2} + \alpha , 0)
    \label{triplet}
\end{equation}
where $\alpha$ is the desired margin between the norm of the anchor $S_{a}$ and the best positive $S_{p}$ and that of $S_{a}$ and the hardest negatives $S^{k}_{n}$ in the descriptor space.
The $k$ is the number of hard negative samples corresponding to each anchor.
We train our model using a set of reference and query databases. 
For each query, we consider it as an anchor, and its positives and negatives are generated from the reference database, which will be detailed in \Refsec{train}. 
\begin{table}[tbp]
    \renewcommand{\arraystretch}{1.3}
    \small
    \begin{center} 
        \caption{datasets detail. This table specifies the number of images in the dataset used.} 
        \label{datasets}
        \begin{tabular}{cc|cllllll}
        \hline
        \multicolumn{2}{c|}{dataset}                                   & \multicolumn{7}{c}{database / queries} \\ \hline
        \multicolumn{1}{c|}{\multirow{5}{*}{MSLS}}     & Melbourne      & \multicolumn{7}{c}{101827 / 88118}     \\
        \multicolumn{1}{c|}{}                          & Amman         & \multicolumn{7}{c}{953 / 835}          \\
        \multicolumn{1}{c|}{}                          & Boston        & \multicolumn{7}{c}{14024 / 6724}       \\
        \multicolumn{1}{c|}{}                          & San Francisco & \multicolumn{7}{c}{6315 / 4525}        \\
        \multicolumn{1}{c|}{}                          & Copenhagen    & \multicolumn{7}{c}{12601 / 6595}       \\ \hline
        \multicolumn{1}{c|}{\multirow{2}{*}{NordLand}} & train set     & \multicolumn{7}{c}{15000 / 15000}      \\
        \multicolumn{1}{c|}{}                          & test set      & \multicolumn{7}{c}{3000 / 3000}        \\ \hline
        \multicolumn{1}{c|}{\multirow{2}{*}{Oxford1}}  & train set     & \multicolumn{7}{c}{2401 / 2448}        \\
        \multicolumn{1}{c|}{}                          & test set      & \multicolumn{7}{c}{1460 / 1474}        \\ \hline
        \multicolumn{1}{c|}{\multirow{2}{*}{Oxford2}}  & train set     & \multicolumn{7}{c}{3619 / 3926}        \\
        \multicolumn{1}{c|}{}                          & test set      & \multicolumn{7}{c}{3632 / 3921}        \\ \hline
        \end{tabular}
    \end{center}
    \vspace{-0.3cm}
\end{table}

\section{EXPERIMENTS}
\subsection{Datasets} \label{dataset}
In our experiments, we use three datasets: MSLS \cite{msls}, NordLand \cite{NordLand}, Oxford RobotCar \cite{OxfordRobotCar}, as summarized in \Reftab{datasets}.
\subsubsection{Mapillary Street Level Sequences (MSLS)}
MSLS is a comprehensive dataset consisting of street-level view image sequences, designed to support VPR studies.
These sequences are collected from various cities.
We used Melbourne for training and Amman, Boston, San Francisco and Copenhagen for testing.
\subsubsection{NordLand}
The Nordland dataset comprises a collection of images captured during of rail journeys across four seasons, covering various weather and lighting conditions.
We use the Summer-Winter pair for training, and Spring-Fall pair for testing.
\subsubsection{Oxford RobotCar}
The Oxford RobotCar dataset is a large-scale dataset for autonomous driving research. 
It encompasses road scenes captured during different time periods.
We design two experimental sub-datasets: Oxford1 and Oxford2.
For Oxford1, we split the database (2015-03-17-11-08-44, day) and query (2014-12-16-18-44-24, night) to train set and test set. For Oxford2, we use a database (2014-12-16-09-14-09, day) and query (2014-12-17-18-18-43, night) for train and database (2014-11-18-13-20-12, day) and query (2014-12-16-18-44-24, night) for test.
These datasets are pre-processed to keep an approximate 2 meters frame separation based on the latitude and longitude of each frame location.
\subsection{ Implementation Details} \label{train}
\begin{table*}[tbp]
    \renewcommand{\arraystretch}{1.4}
    \begin{center}
        \caption{Quantitative results on MSLS} 
        \label{tb_quantitative1}
        \setlength{\tabcolsep}{1.9mm}{
        \begin{tabular}{c|c|cccccccccccc}
        \hline
        \multirow{3}{*}{Method} &\multirow{3}{*}{Dimension} &\multicolumn{12}{c}{MSLS}                                                                                                                                      \\ \cline{3-14} 
                                &               & \multicolumn{3}{c|}{Amman}                 & \multicolumn{3}{c|}{Boston}                & \multicolumn{3}{c|}{SF}                    & \multicolumn{3}{c}{Cph} \\ \cline{3-14} 
                                &                    & R@1   & R@5   & \multicolumn{1}{c|}{R@10}  & R@1   & R@5   & \multicolumn{1}{c|}{R@10}  & R@1   & R@5   & \multicolumn{1}{c|}{R@10}  & R@1    & R@5    & R@10  \\ \hline
        NetVLAD \cite{arandjelovic2016netvlad} &4096 & 0.189 & 0.251 & \multicolumn{1}{c|}{0.277} & 0.179 & 0.238 & \multicolumn{1}{c|}{0.267} & 0.289 & 0.398 & \multicolumn{1}{c|}{0.455} & 0.405 & 0.534 & 0.594 \\
        NetVLAD+SeqMatch \cite{garg2021seqnet} &4096 & 0.246 & 0.302 & \multicolumn{1}{c|}{0.330} & 0.204 & 0.239 & \multicolumn{1}{c|}{0.257} & 0.363 & 0.430 & \multicolumn{1}{c|}{0.460} & 0.504 & 0.612 & 0.657 \\
        SeqNet \cite{garg2021seqnet}           &4096 & 0.269 & 0.376 & \multicolumn{1}{c|}{0.408} & 0.274 & 0.354 & \multicolumn{1}{c|}{0.390} & 0.556 & 0.671 & \multicolumn{1}{c|}{0.728} & 0.462 & 0.581 & 0.637 \\
        SeqVLAD \cite{mereu2022seqvlad}        &24576 & 0.300 & \textbf{0.448} & \multicolumn{1}{c|}{\underline{0.519}} & 0.466 & 0.628 & \multicolumn{1}{c|}{0.678} & 0.661 & 0.826 & \multicolumn{1}{c|}{\underline{0.863}} & 0.564 & 0.722 & 0.777 \\
        Ours                                   &24576 & \underline{0.303} & 0.423 & \multicolumn{1}{c|}{0.511} & \textbf{0.504} & \textbf{0.645} & \multicolumn{1}{c|}{\underline{0.688}} & \textbf{0.680} & \textbf{0.841} & \multicolumn{1}{c|}{\textbf{0.864}} & \textbf{0.608} & \textbf{0.765} & \textbf{0.801} \\ \hline
        SeqVLAD w/ PCA                         &4096 & 0.294 & \underline{0.442} & \multicolumn{1}{c|}{\textbf{0.526}} & 0.465 & 0.623 & \multicolumn{1}{c|}{0.675} & 0.656 & 0.822 & \multicolumn{1}{c|}{0.859} & 0.560 & 0.720 & 0.774 \\
        Ours w/ PCA                            &4096 & \textbf{0.306} & 0.411 & \multicolumn{1}{c|}{0.510} & \underline{0.502} & \textbf{0.645} & \multicolumn{1}{c|}{\textbf{0.691}} & \underline{0.671} & \underline{0.839} & \multicolumn{1}{c|}{0.860} & \underline{0.604} & \underline{0.760} & \textbf{0.801} \\ \hline
        \end{tabular}
        }
    \end{center}
    The best and second-best results for each dataset are highlighted. The best overall results on each dataset are indicated in \textbf{bold}, while the second-best results are \underline{underlined}.
    \vspace{-0.3cm}
\end{table*}
\begin{table*}[tbp]
    \renewcommand{\arraystretch}{1.4}
    \begin{center}
        \caption{Quantitative results on Nordland and Oxford RobotCar} 
        \label{tb_quantitative2}
        \setlength{\tabcolsep}{1.9mm}{
        \begin{tabular}{c|c|ccc|ccc|ccc}
        \hline
        \multirow{2}{*}{Method} &\multirow{2}{*}{Dimension} & \multicolumn{3}{c|}{NordLand} & \multicolumn{3}{c|}{Oxford1} & \multicolumn{3}{c}{Oxford2} \\ \cline{3-11} 
                                &                      & R@1      & R@5      & R@10    & R@1      & R@5     & R@10    & R@1     & R@5     & R@10    \\ \hline
        NetVLAD \cite{arandjelovic2016netvlad} &4096        & 0.377    & 0.543    & 0.615   & 0.468    & 0.696   & 0.779   & -       & -       & -       \\
        NetVLAD+SeqMatch \cite{garg2021seqnet} &4096       & 0.610    & 0.705    & 0.746   & 0.672    & 0.784   & 0.846   & -       & -       & -       \\
        SeqNet \cite{garg2021seqnet}           &4096       & 0.797    & 0.905    & 0.930   & 0.741    & 0.875   & 0.933   & -       & -       & -       \\
        SeqVLAD \cite{mereu2022seqvlad}        &24576      & 0.964    & 0.992    & 0.993   & \underline{0.966}    & \underline{0.982}   & \underline{0.989}   & 0.844   & 0.929   & 0.958   \\
        Ours                                   &24576      & \textbf{0.971}    & \textbf{0.995}    & \textbf{0.995}   & 0.958    & 0.978   & 0.988   & \textbf{0.868}   & \underline{0.944}   & \underline{0.968}  \\ \hline
        SeqVLAD w/ PCA                         &4096       & 0.963    & 0.991    & 0.994   & \textbf{0.967}    & \textbf{0.982}   & \textbf{0.990}   & 0.847   & 0.932   & 0.961   \\                            
        Ours w/ PCA                            &4096       & \textbf{0.971}    & \textbf{0.995}    & \textbf{0.995}   & 0.955    & 0.977   & 0.986   & \underline{0.866}   & \textbf{0.945}   & \textbf{0.969}   \\ \hline  
        \end{tabular}
        }
    \end{center}
    The best and second-best results for each dataset are highlighted. The best overall results on each dataset are indicated in \textbf{bold}, while the second-best results are \underline{underlined}. 
    \vspace{-0.3cm}
\end{table*}
\textbf{Architecture.} We implement our method using the Pytorch framework \cite{paszke2019pytorch} on an NVIDIA RTX A6000 card.
In the patch embedding process, the CNN comprises two convolutional layers.
The first layer maps 3 channels to 64 channels and the second layer maps 64 channels to 384 channels.
We set the convolution parameters as follows: $kernel=7$, $stride = 2$ and $padding = 1$.
After each convolution operation, we apply the ReLU activation function followed by max pooling.
Then we incorporate the spatial transformer encoder with $L_{s}=4$ layers and the temporal transformer encoder with $L_{t}=4$ layers.
Additionally, we use a multi-head in transformer with $h=6$ heads.
In the temporal transformer encoder, we set the size of sliding window $m=6$ and the $stride=3$.
Both the inputs and outputs of the transformers are embeddings with a dimensionality of $D=384$.
In the NetVLAD module, we configure the number of clusters to be 64, yielding sequence descriptors with dimensions of 384 $\times$ 64 without the application of Principal Component Analysis (PCA) \cite{jegou2012negativepca}.
To facilitate comparison with other methods, we perform dimensionality reduction using PCA, reducing the dimensionality to 4096.

\textbf{Training.} In the training phase, we initialize the model with pre-trained parameters from CCT \cite{hassani2021cct} and adopt the Adam optimizer \cite{kingma2014adam}.
All images are resized to 384 $\times$ 384.
The learning rates for spatial transformer encoder, temporal transformer encoder and NetVLAD are configured as $0.0001$, $0.001$ and $0.0001$ respectively.
We set the $batch\ size = 4$, with each batch consisting of a query sequence, a best positive sequence and 5 hardest negative sequences ($k=5$ as referenced in \Refeq{triplet}).
The length of each sequence is set to $L=5$.
The margin in triplet loss is specified as $\alpha=0.1$.
The mining method \cite{arandjelovic2016netvlad} is used to select samples, \ie we initially select samples based on GNSS labels between the query and the database, and then we select the best positive and the hardest negatives by cosine distance in the descriptor space.
Since selecting negatives from the whole dataset is time and space consuming, we adopt partial mining \cite{msls}.
This involves randomly sampling a subset of negatives using GNSS labels filtering and using a cache to store the descriptors of sub-negatives.
The cache is employed for selecting negatives and is refreshed after every 1000 iterations.
We implement early stopping by halting the training if the Recall@5 does improve for 5 consecutive epochs.
We set the positive distance threshold to 10 meters and the negative distance threshold to 25 meters.

\textbf{Evaluation.} In the evaluation phase, we use Recall@K as the performance metric.
Recall@K is defined as the ratio of the number of correct queries retrieved to the total number of queries.
A correct retrieval is defined as at least one of the top $K$ retrieval being within the given radius from the ground truth position of the query.
We use radii of 10 meters, 20 meters and 1 frame for Oxford, MSLS and Nordland datasets respectively.

\subsection{Results}
\subsubsection{Comparison with the State-of-the-art Methods}
The chosen baseline methods include the state-of-the-art methods using sequence descriptors, \ie SeqNet \cite{garg2021seqnet} and SeqVLAD \cite{mereu2022seqvlad}.
Additionally, we also compare our method with NetVLAD \cite{arandjelovic2016netvlad} and NetVLAD+SeqMatch \cite{garg2021seqnet}.
To ensure a fair comparison, all the experimental results are reproduced via our setting described in \Refsec{dataset} and \Refsec{train}.

\Reftab{tb_quantitative1} and \Reftab{tb_quantitative2} show the results of our method compared to the baseline methods on MSLS, NordLand and Oxford RobotCar.
It is evident that NetVLAD performs worse than other methods, indicating that sequence VPR significantly outperforms the single-frame VPR.
In addition, the methods based on sequence descriptors outperform the method based on SeqMatch.

Compared to SeqNet, our method outperforms it across all datasets. We observe a Recall@1 improvement of over 10\% in most datasets, except for a 4\% improvement in Amman.
SeqNet generates a sequence descriptor through a weighted sum of frame descriptors, which are created by aggregating the local features of each frame. While local features can be discriminative for individual frames within a sequence, they may not exhibit the same level of discriminative power across all sequences.
In contrast, our method directly derives the sequence descriptors from the local features of all frames within a sequence. This approach ensures that our descriptor maintains its discriminative qualities across different sequences.

\begin{figure}
    \centering
    \includegraphics[width=3.4in]{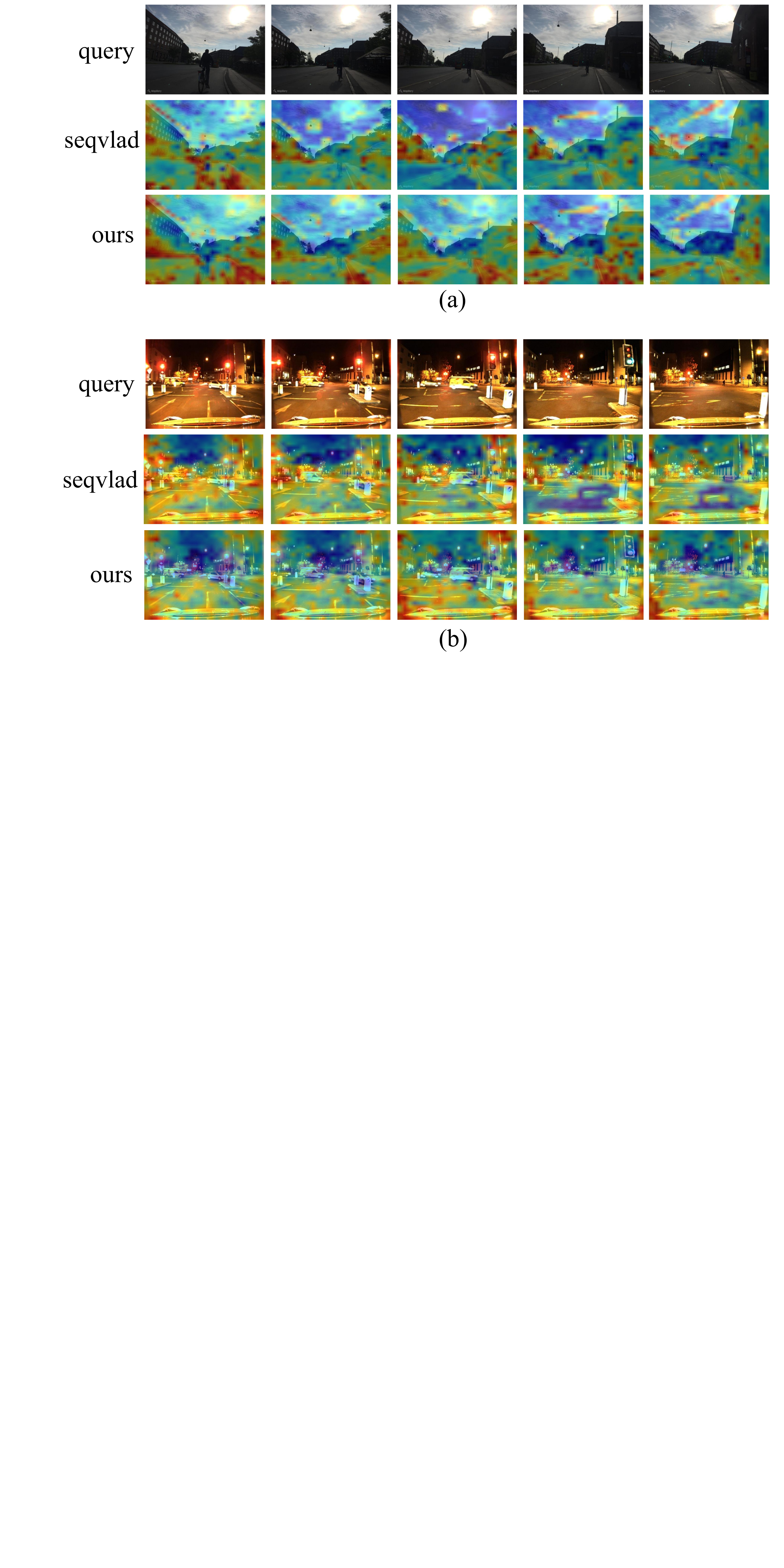}
    \caption{\textbf{Visualizations on attention.}
    Here are the attentions of our method and SeqVLAD for different regions of the query sequences which is in \Reffig{qualitative} (a), \Reffig{qualitative} (c).
    Red portions indicate more focus, and blue portions indicate less focus.
    Compared to SeqVLAD, our method focuses less on dynamic objects and more on road elements.
    }
    \label{attention_map}
\end{figure}

Additionally, compared with SeqVLAD, our method exhibits superior performance in most datasets, except Amman and Oxford1.
Notably, SeqVLAD does not take into account the temporal correlation across multiple frames.
As shown in \Reffig{qualitative} (a)(c), the SeqVLAD is susceptible to dynamic objects, \eg{bicyclists}, and is sensitive to illumination changes from day to night or variations in weather conditions.
Conversely, our proposed cross-frames temporal attention can effectively capture local regions correspondences to learn patterns that persist over time.
This property renders our sequence descriptors more robust to illumination changes and local scene variations.
\Reffig{attention_map} provides further insight by illustrating the attention mechanisms of both our method and SeqVLAD for different regions within query sequences from \Reffig{qualitative} (a)(c), substantiating the aforementioned conclusions.
While our method's performance in Oxford1 is slightly lower than SeqVLAD, there is a noteworthy Recall@1 improvement of over 2\% in Oxford2.
Oxford1 has a smaller train set compared to Oxford2, but our model has a higher parameter count than SeqVLAD, making it more challenging to train effectively.
On the other hand, SeqVLAD tends to be more susceptible to overfitting.

Finally, we delve into the impact of reducing the dimensionality of sequence descriptors.
The results reveal that when descriptors undergo dimensionality reduction via Principal Component Analysis (PCA) into a 4096-Dimensional space, their performance remains on par with that of the full-sized descriptors.

\begin{figure}
    \centering
    \includegraphics[width=3in]{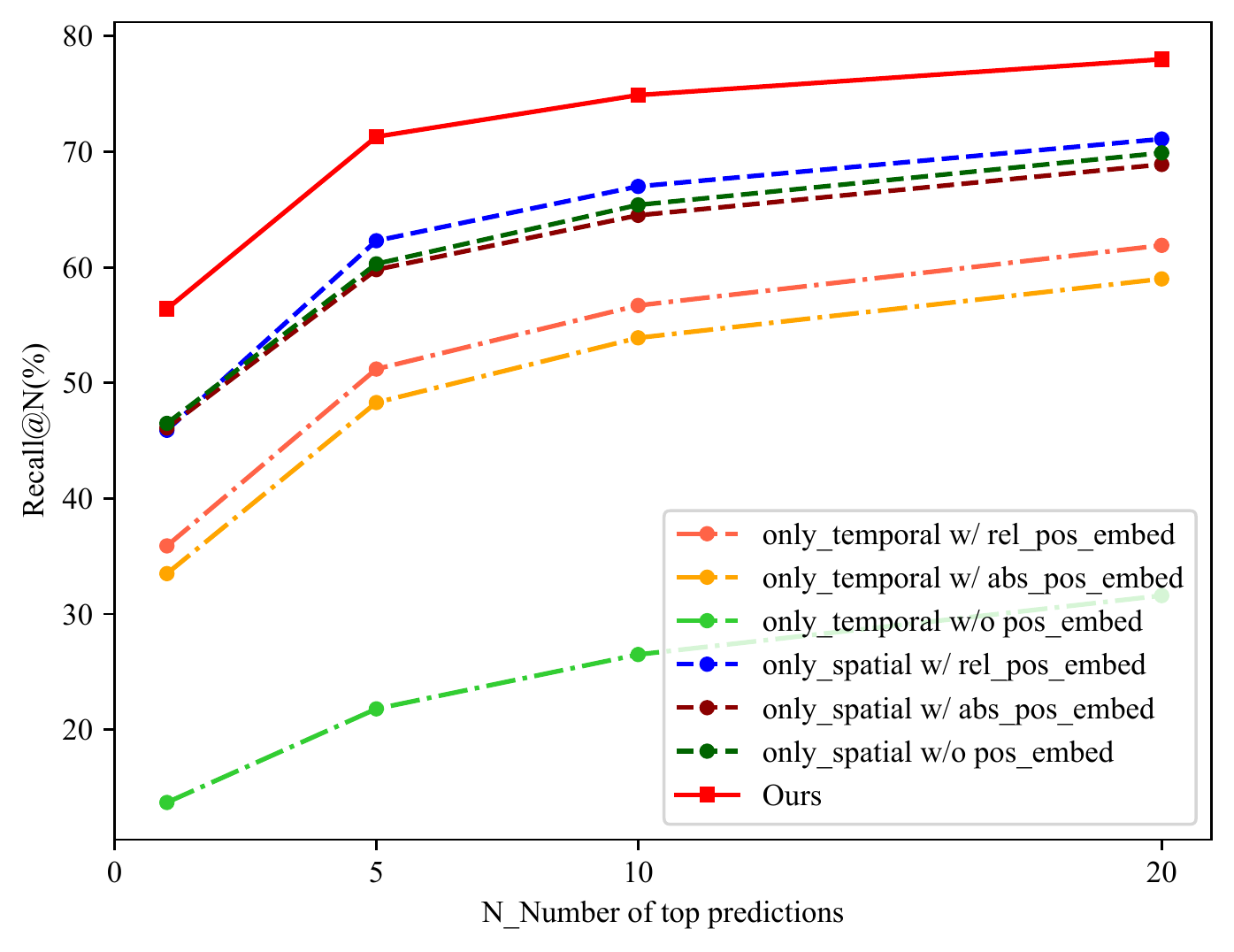}
    \caption{\textbf{Ablation Studies for spatio-temporal effectiveness and positional embedding.}
    We show the comparison of Recall@N performances with only spatial or temporal module, and two kinds of positional embedding or w/o position information.
    In addition, the positional embedding is trained from scratch without pre-trained parameters.}
    \label{ablation1}
\end{figure}

\subsubsection{Ablation Studies}
We conduct ablation studies on the four test cities of MSLS to analyze the effectiveness of the \emph{\textbf{spatio-temporal attention with positional embedding}} and the \emph{\textbf{sliding window setting}}.
As shown in \Reffig{ablation1}, we compare the experimental results of spatio-temporal attention with positional embedding.
It can be clearly observed that descriptors extracted via spatial attention achieve better performance than those extracted via temporal attention.
This may indicate that the spatial structure plays a dominant role in the sequence descriptors.
But our sequence descriptors extracted via spatio-temporal attention achieve the best performance.
This suggests that fusing temporal information to spatial structure can further improve the representation of the descriptors.
Furthermore, the role of positional embedding is slight for spatial attention but crucial for temporal attention.
Fusing position information can greatly improve performance of descriptors, and relative positional embedding is superior to absolute positional embedding.

\begin{figure}
    \centering
    \includegraphics[width=3in]{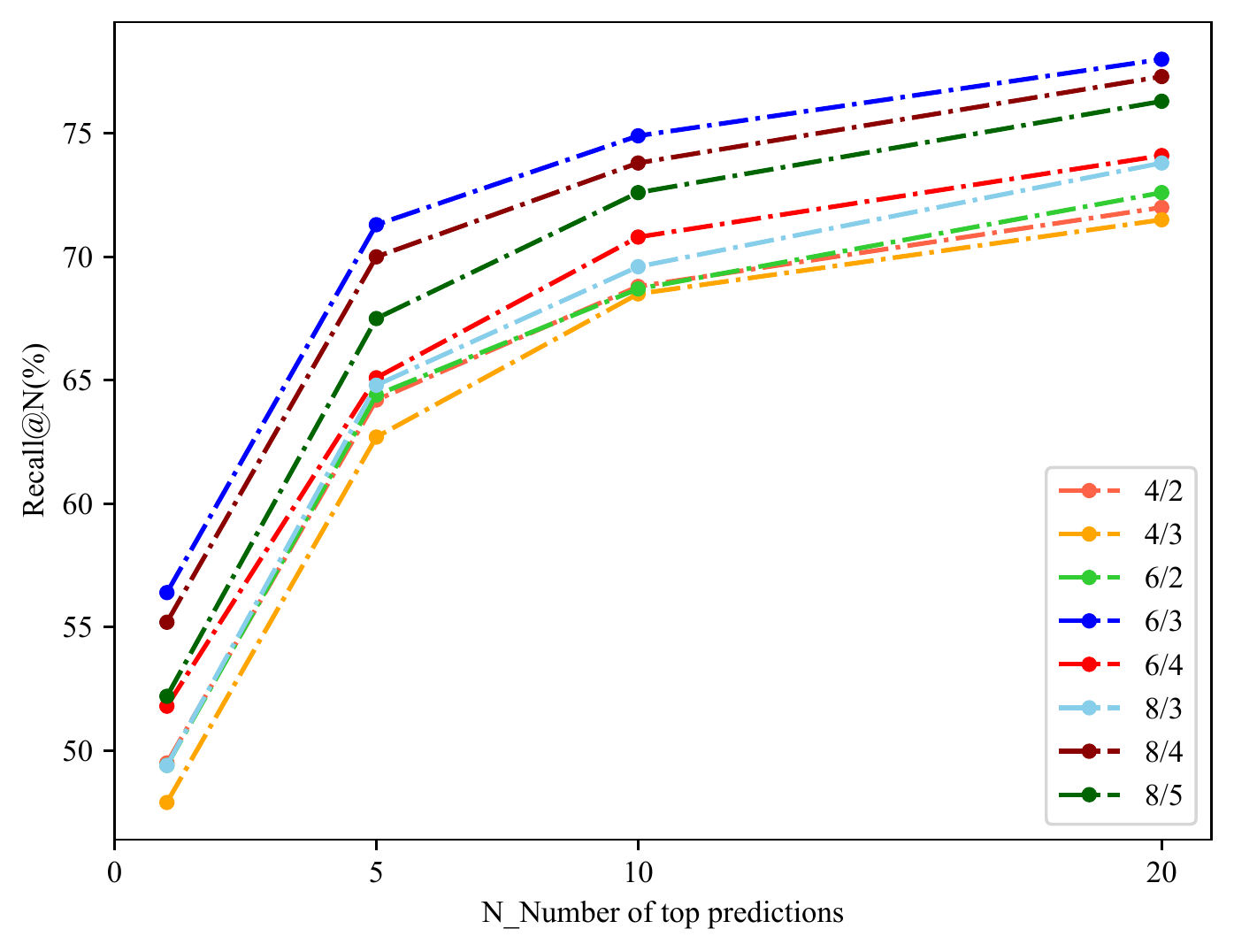}
    \caption{\textbf{Ablation Studies for sliding window settings.}
    We show the comparison of Recall@N performances with different sliding window settings, where m/s demonstrates the size and stride of the sliding window respectively.
    }
    \label{ablation2}
\end{figure}
In addition, we explore how the hypermeters of the sliding window affect the ability to capture the dynamics of local features in temporal attention.
We compare different sliding window settings, where $4\times 4$, $6\times 6$ and $8\times 8$ are set for the size of window, and \{2, 3\}, \{2, 3, 4\} and \{3, 4, 5\} are set for the stride respectively.
As observed in the \Reffig{ablation2}, for a given stride, a larger sliding window performs better, but the performance decreases beyond a certain threshold.
Finally, the optimal value of stride is half of the sliding window size.

\subsubsection{Runtime Analysis and Memory Footprint}
\begin{table}[]
    \renewcommand{\arraystretch}{1.4}
    \caption{Resource consumption and model size. The ms/fra indicates the time of extracting a frame descriptor, and ms/seq indicate the time of extracting a sequence descriptor.}
    \label{consumption}
    \begin{tabular}{c|c|c|c|c}
    \hline
    Method  & \begin{tabular}[c]{@{}c@{}}Extraction \\ latency\end{tabular} & \begin{tabular}[c]{@{}c@{}}GPU \\ Memory \end{tabular} & GFLOPs & Params  \\ \hline
    NetVLAD \cite{arandjelovic2016netvlad} & 8.8 ms/fra                                                  & 57.26 MB                                                    & 45.12  & 14.74 M     \\
    SeqNet \cite{garg2021seqnet}  & 6.2 ms/seq                                            & 320.01 MB                                                    & 0.84   & 83.89 M     \\
    SeqVLAD \cite{mereu2022seqvlad} & 8.9 ms/seq                                             & 59.88 MB                                                     & 32.71  & 7.15 M      \\
    Ours     & 18.7 ms/seq                                           & 103.74 MB                                                    & 63.74  & 13.06 M      \\ \hline
    \end{tabular}
\end{table}
\begin{figure}
    \centering
    \includegraphics[height=15.5cm]{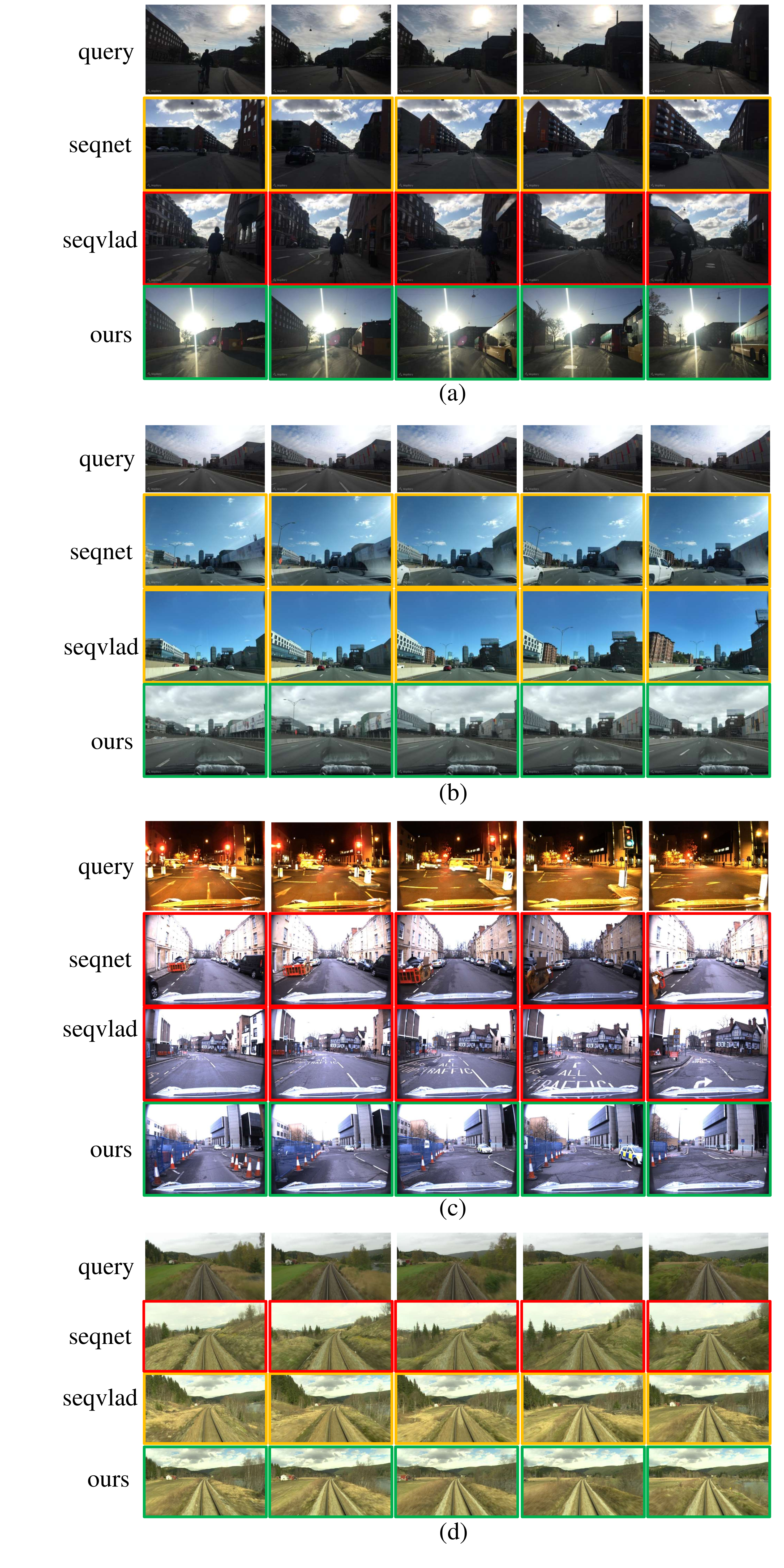}
    \caption{\textbf{Qualitative results.}
    In these examples, the proposed method successfully retrieves the matching reference sequence in MSLS street (a) view and highway (b), Oxford (c) and NordLand (d), while SeqNet and SeqVLAD produce incorrect place matches.
    Green and red indicate correct and incorrect retrievals, respectively. While orange indicates the same view but beyond a certain GNSS label threshold, which is also defined as incorrect retrievals.
    }
    \label{qualitative}
\end{figure}
In real-world VPR systems, it is crucial to take latency and scalability into account.
\Reftab{consumption} provides insights into the computational time, GPU memory footprint and model size of the compared techniques in evaluation.
SeqNet is able to extract sequence descriptors more swiftly and with lower GFLOPs due to the pre-extraction and storage of NetVLAD descriptors for each image offline. This eliminates the need to account for the time taken by NetVLAD. 
In addition, the memory footprint and model parameters of SeqNet are influenced by both the descriptor dimension and sequence length, which are proportional to $D \times D \times L$, where $D$ represents the descriptor dimension and $L$ represents the sequence length.
In contrast, our approach and SeqVLAD extract sequence descriptors directly from the original image sequences, with the entire process being executed online.
Additionally, our model considers the interaction among consecutive frames, making it more intricate than SeqVLAD.
Consequently, it takes more time to extract a sequence descriptor.

\subsubsection{Qualitative Results}

In \Reffig{qualitative}, we show our retrieval sequence compared with those from SeqNet and SeqVLAD in MSLS street view and highway, Oxford and NordLand.
Sequences marked with green and red borders indicate correct and incorrect retrievals, respectively.
Sequences marked with orange borders indicate that the retrieval sequence and the query have the same view, but their GNSS labels define that they are not the same ``place''.
Based on the qualitative results, our method demonstrates the capability to handle changes in lighting conditions caused by day-night transitions and weather changes.
In addition, it is less susceptible to dynamic occlusions and partial scene changes, such as pedestrians and vehicles on the road, as well as changes due to road maintenance.

\subsection{Limitations}
The complexity of our model results in a heightened reliance on the size and distribution of the train set, which may yield subpar results when the train set is small, as observed in \Reftab{tb_quantitative2} Oxford1.
Additionally, we further analyze the failure cases in Amman. We find that some query images and their ground truth exhibit a large discrepancy in field of view (FOV). Consequently, our approach, which incorporates temporal interaction, may introduce greater temporal consistency in sequence descriptors than descriptors without temporal information. This could be one of the factors contributing to the failure cases.

\section{CONCLUSION}
VPR holds immense potential for various applications.
Our work aims to provide a new perspective on sequence-based VPR.
Instead of aggregating multiple frames spatially, we introduce the fusion of features in the temporal dimension.
We use a spatio-temporal attention approach to generate a discriminative descriptor of sequences with improved accuracy compared to existing methods.
Additionally, our findings emphasize the significance of both spatial structure and temporal variation in sequence descriptors.
We anticipate that these insights will serve as a solid foundation for future research endeavors, enabling improved utilization of sequence information in VPR.

\bibliographystyle{IEEEtran}
\bibliography{IEEEabrv, paper}

\begin{thebibliography}{10}
\providecommand{\url}[1]{#1}
\csname url@rmstyle\endcsname
\providecommand{\newblock}{\relax}
\providecommand{\bibinfo}[2]{#2}
\providecommand\BIBentrySTDinterwordspacing{\spaceskip=0pt\relax}
\providecommand\BIBentryALTinterwordstretchfactor{4}
\providecommand\BIBentryALTinterwordspacing{\spaceskip=\fontdimen2\font plus
\BIBentryALTinterwordstretchfactor\fontdimen3\font minus
  \fontdimen4\font\relax}
\providecommand\BIBforeignlanguage[2]{{%
\expandafter\ifx\csname l@#1\endcsname\relax
\typeout{** WARNING: IEEEtran.bst: No hyphenation pattern has been}%
\typeout{** loaded for the language `#1'. Using the pattern for}%
\typeout{** the default language instead.}%
\else
\language=\csname l@#1\endcsname
\fi
#2}}

\bibitem{lowry2015survey}
S.~Lowry, N.~S{\"u}nderhauf, P.~Newman, J.~J. Leonard, D.~Cox, P.~Corke, and
  M.~J. Milford, ``Visual place recognition: A survey,'' \emph{ieee
  transactions on robotics}, vol.~32, no.~1, pp. 1--19, 2015.

\bibitem{vlad}
H.~Jégou, M.~Douze, C.~Schmid, and P.~Pérez, ``Aggregating local descriptors
  into a compact image representation,'' in \emph{2010 IEEE Computer Society
  Conference on Computer Vision and Pattern Recognition}, 2010, pp. 3304--3311.

\bibitem{fishervector}
J.~S{\'a}nchez, F.~Perronnin, T.~Mensink, and J.~Verbeek, ``Image
  classification with the fisher vector: Theory and practice,''
  \emph{International journal of computer vision}, vol. 105, no.~3, pp.
  222--245, 2013.

\bibitem{bow}
H.~J{\'e}gou, F.~Perronnin, M.~Douze, J.~S{\'a}nchez, P.~P{\'e}rez, and
  C.~Schmid, ``Aggregating local image descriptors into compact codes,''
  \emph{IEEE transactions on pattern analysis and machine intelligence},
  vol.~34, no.~9, pp. 1704--1716, 2011.

\bibitem{chen2014convolutional}
Z.~Chen, O.~Lam, A.~Jacobson, and M.~Milford, ``Convolutional neural
  network-based place recognition,'' in \emph{Proceedings of the 16th
  Australasian Conference on Robotics and Automation 2014}.\hskip 1em plus
  0.5em minus 0.4em\relax Australian Robotics and Automation Association
  (ARAA), 2014, pp. 1--8.

\bibitem{arandjelovic2016netvlad}
R.~Arandjelovic, P.~Gronat, A.~Torii, T.~Pajdla, and J.~Sivic, ``Netvlad: Cnn
  architecture for weakly supervised place recognition,'' in \emph{Proceedings
  of the IEEE conference on computer vision and pattern recognition}, 2016, pp.
  5297--5307.

\bibitem{jin2017crn}
H.~Jin~Kim, E.~Dunn, and J.-M. Frahm, ``Learned contextual feature reweighting
  for image geo-localization,'' in \emph{Proceedings of the IEEE Conference on
  Computer Vision and Pattern Recognition}, 2017, pp. 2136--2145.

\bibitem{milford2012seqslam}
M.~J. Milford and G.~F. Wyeth, ``Seqslam: Visual route-based navigation for
  sunny summer days and stormy winter nights,'' in \emph{2012 IEEE
  international conference on robotics and automation}.\hskip 1em plus 0.5em
  minus 0.4em\relax IEEE, 2012, pp. 1643--1649.

\bibitem{siam2017fastseqslam}
S.~M. Siam and H.~Zhang, ``Fast-seqslam: A fast appearance based place
  recognition algorithm,'' in \emph{2017 IEEE International Conference on
  Robotics and Automation (ICRA)}.\hskip 1em plus 0.5em minus 0.4em\relax IEEE,
  2017, pp. 5702--5708.

\bibitem{facil2019condition}
J.~M. Facil, D.~Olid, L.~Montesano, and J.~Civera, ``Condition-invariant
  multi-view place recognition,'' \emph{arXiv preprint arXiv:1902.09516}, 2019.

\bibitem{garg2021seqnet}
S.~Garg and M.~Milford, ``Seqnet: Learning descriptors for sequence-based
  hierarchical place recognition,'' \emph{IEEE Robotics and Automation
  Letters}, vol.~6, no.~3, pp. 4305--4312, 2021.

\bibitem{mereu2022seqvlad}
R.~Mereu, G.~Trivigno, G.~Berton, C.~Masone, and B.~Caputo, ``Learning
  sequential descriptors for sequence-based visual place recognition,''
  \emph{IEEE Robotics and Automation Letters}, vol.~7, no.~4, pp.
  10\,383--10\,390, 2022.

\bibitem{lu2019dtw}
F.~Lu, B.~Chen, Z.~Guo, and X.~Zhou, ``Visual sequence place recognition with
  improved dynamic time warping,'' in \emph{2019 IEEE 31st International
  Conference on Tools with Artificial Intelligence (ICTAI)}.\hskip 1em plus
  0.5em minus 0.4em\relax IEEE, 2019, pp. 1034--1041.

\bibitem{naseer2014robust}
T.~Naseer, L.~Spinello, W.~Burgard, and C.~Stachniss, ``Robust visual robot
  localization across seasons using network flows,'' in \emph{Proceedings of
  the AAAI conference on artificial intelligence}, vol.~28, no.~1, 2014.

\bibitem{vysotska2019effective}
O.~Vysotska and C.~Stachniss, ``Effective visual place recognition using
  multi-sequence maps,'' \emph{IEEE Robotics and Automation Letters}, vol.~4,
  no.~2, pp. 1730--1736, 2019.

\bibitem{feichtenhofer2017spatiotemporal}
C.~Feichtenhofer, A.~Pinz, and R.~P. Wildes, ``Spatiotemporal multiplier
  networks for video action recognition,'' in \emph{Proceedings of the IEEE
  conference on computer vision and pattern recognition}, 2017, pp. 4768--4777.

\bibitem{li2020spatio}
J.~Li, X.~Liu, W.~Zhang, M.~Zhang, J.~Song, and N.~Sebe, ``Spatio-temporal
  attention networks for action recognition and detection,'' \emph{IEEE
  Transactions on Multimedia}, vol.~22, no.~11, pp. 2990--3001, 2020.

\bibitem{bertasius2021space}
G.~Bertasius, H.~Wang, and L.~Torresani, ``Is space-time attention all you need
  for video understanding?'' in \emph{ICML}, vol.~2, no.~3, 2021, p.~4.

\bibitem{li2022mvitv2}
Y.~Li, C.-Y. Wu, H.~Fan, K.~Mangalam, B.~Xiong, J.~Malik, and C.~Feichtenhofer,
  ``Mvitv2: Improved multiscale vision transformers for classification and
  detection,'' in \emph{Proceedings of the IEEE/CVF Conference on Computer
  Vision and Pattern Recognition}, 2022, pp. 4804--4814.

\bibitem{ji2020spatio}
J.~Ji, C.~Xu, X.~Zhang, B.~Wang, and X.~Song, ``Spatio-temporal memory
  attention for image captioning,'' \emph{IEEE Transactions on Image
  Processing}, vol.~29, pp. 7615--7628, 2020.

\bibitem{aich2021spatio}
A.~Aich, M.~Zheng, S.~Karanam, T.~Chen, A.~K. Roy-Chowdhury, and Z.~Wu,
  ``Spatio-temporal representation factorization for video-based person
  re-identification,'' in \emph{Proceedings of the IEEE/CVF international
  conference on computer vision}, 2021, pp. 152--162.

\bibitem{dosovitskiy2021anvit}
A.~Dosovitskiy, L.~Beyer, A.~Kolesnikov, D.~Weissenborn, X.~Zhai,
  T.~Unterthiner, M.~Dehghani, M.~Minderer, G.~Heigold, S.~Gelly, J.~Uszkoreit,
  and N.~Houlsby, ``An image is worth 16x16 words: Transformers for image
  recognition at scale,'' in \emph{International Conference on Learning
  Representations}, 2021.

\bibitem{hassani2021cct}
A.~Hassani, S.~Walton, N.~Shah, A.~Abuduweili, J.~Li, and H.~Shi, ``Escaping
  the big data paradigm with compact transformers,'' \emph{arXiv preprint
  arXiv:2104.05704}, 2021.

\bibitem{shaw2018self}
P.~Shaw, J.~Uszkoreit, and A.~Vaswani, ``Self-attention with relative position
  representations,'' in \emph{Proceedings of the 2018 Conference of the North
  American Chapter of the Association for Computational Linguistics: Human
  Language Technologies, Volume 2 (Short Papers)}, 2018, pp. 464--468.

\bibitem{msls}
F.~Warburg, S.~Hauberg, M.~Lopez-Antequera, P.~Gargallo, Y.~Kuang, and
  J.~Civera, ``Mapillary street-level sequences: A dataset for lifelong place
  recognition,'' in \emph{Proceedings of the IEEE/CVF conference on computer
  vision and pattern recognition}, 2020, pp. 2626--2635.

\bibitem{NordLand}
N.~S{\"u}nderhauf, P.~Neubert, and P.~Protzel, ``Are we there yet? challenging
  seqslam on a 3000 km journey across all four seasons,'' in \emph{Proc. of
  workshop on long-term autonomy, IEEE international conference on robotics and
  automation (ICRA)}, 2013, p. 2013.

\bibitem{OxfordRobotCar}
W.~Maddern, G.~Pascoe, C.~Linegar, and P.~Newman, ``1 year, 1000 km: The oxford
  robotcar dataset,'' \emph{The International Journal of Robotics Research},
  vol.~36, no.~1, pp. 3--15, 2017.

\bibitem{paszke2019pytorch}
A.~Paszke, S.~Gross, F.~Massa, A.~Lerer, J.~Bradbury, G.~Chanan, T.~Killeen,
  Z.~Lin, N.~Gimelshein, L.~Antiga, \emph{et~al.}, ``Pytorch: An imperative
  style, high-performance deep learning library,'' \emph{Advances in neural
  information processing systems}, vol.~32, pp. 8026--8037, 2019.

\bibitem{jegou2012negativepca}
H.~J{\'e}gou and O.~Chum, ``Negative evidences and co-occurences in image
  retrieval: The benefit of pca and whitening,'' in \emph{Computer Vision--ECCV
  2012: 12th European Conference on Computer Vision, Florence, Italy, October
  7-13, 2012, Proceedings, Part II 12}.\hskip 1em plus 0.5em minus 0.4em\relax
  Springer, 2012, pp. 774--787.

\bibitem{kingma2014adam}
D.~P. Kingma and J.~Ba, ``Adam: a method for stochastic optimization,'' in
  \emph{International Conference on Learning Representations}, 2015, pp. 1--15.

\end{thebibliography}

\end{document}